\documentclass[conference]{IEEEtran}
\IEEEoverridecommandlockouts
\usepackage{cite}
\usepackage{amsmath,amssymb,amsfonts}
\usepackage{algorithmic}
\usepackage{graphicx}
\usepackage{textcomp}
\usepackage{xcolor}
\def\BibTeX{{\rm B\kern-.05em{\sc i\kern-.025em b}\kern-.08em
    T\kern-.1667em\lower.7ex\hbox{E}\kern-.125emX}}
    
\usepackage{booktabs}
\usepackage{multirow}
\usepackage{subfig}
\usepackage{hyperref}
    
\makeatletter
\def\endthebibliography{%
  \def\@noitemerr{\@latex@warning{Empty `thebibliography' environment}}%
  \endlist
}    
   
\newcommand{\etal}{\textit{et al.}}

\begin{document}

\title{Augmenting Ego-Vehicle for Traffic Near-Miss and Accident Classification Dataset using Manipulating Conditional Style Translation}

\author{\IEEEauthorblockN{Hilmil Pradana}
\IEEEauthorblockA{\textit{Big Data Integration Research Center} \\
\textit{NICT}\\
Tokyo, Japan \\
hilmi@nict.go.jp}
\and
\IEEEauthorblockN{Minh-Son Dao}
\IEEEauthorblockA{\textit{Big Data Integration Research Center} \\
\textit{NICT}\\
Tokyo, Japan \\
dao@nict.go.jp}
\and
\IEEEauthorblockN{Koji Zettsu}
\IEEEauthorblockA{\textit{Big Data Integration Research Center} \\
\textit{NICT}\\
Tokyo, Japan \\
zettsu@nict.go.jp}
}

\maketitle

\begin{abstract}
To develop the advanced self-driving systems, many researchers are focusing to alert all possible traffic risk cases from closed-circuit television (CCTV) and dashboard-mounted cameras.
Most of these methods focused on identifying frame-by-frame in which an anomaly has occurred, but they are unrealized, which road traffic participant can cause ego-vehicle leading into collision because of available annotation dataset only to detect anomaly on traffic video.
Near-miss is one type of accident and can be defined as a narrowly avoided accident.
However, there is no difference between accident and near-miss at the time before the accident happened, so our contribution is to redefine the accident definition and re-annotate the accident inconsistency on DADA-2000 dataset together with near-miss.
By extending the start and end time of accident duration, our annotation can precisely cover all ego-motions during an incident and consistently classify all possible traffic risk accidents including near-miss to give more critical information for real-world driving assistance systems.
The proposed method integrates two different components: conditional style translation (CST) and separable 3-dimensional convolutional neural network (S3D).
CST architecture is derived by unsupervised image-to-image translation networks (UNIT) used for augmenting the re-annotation DADA-2000 dataset to increase the number of traffic risk accident videos and to generalize the performance of video classification model on different types of conditions while S3D is useful for video classification to prove dataset re-annotation consistency.
In evaluation, the proposed method achieved a significant improvement result by $\textbf{10.25 \%}$ positive margin from the baseline model for accuracy on cross-validation analysis.
Quantitative evaluation based on our re-annotation shows that the proposed method is valuable for computer vision communities to train their models to produce better traffic risk classification.
\end{abstract}


\section{Introduction}
Recently, advanced self-driving car brings significantly improvement technology on various aspects such as efficiency, convenience, and transportation safety system to contribute the global society impacts around the world. 
The potential impact to address self-driving technology is to grow industrial application of automotive driving system \cite{Daily2017, Swief2018, Bocca2019, Karuppasamy2021}.
In most technological breakthroughs, most of researchers play an important role to contribute creative, innovative, and novel ideas, proven concept systems, and continuously improving the advancing technologies to realize autonomous driving in reality.
The goal of self-driving car is to guarantee a safe driving strategy using automatic intelligent systems \cite{Bocca2019}.
Among intelligent system techniques, computer vision algorithms have highly contributed to detect all potential risks in real application.

On this decade, computer vision together with artificial intelligence has realized self-driving system on modern vehicles.
It is started from smart driving assistance that can monitor the surrounding objects and support the driver to alert them in case of an emergency or accident risk.
To develop smart driving assistance, many researchers are focusing to alert all possible traffic risk cases from closed-circuit television (CCTV) \cite{Huang2020, Robles2021, Pawar2021} and dashboard-mounted cameras \cite{Suzuki2018, yao2019, Harest2020, bao2020, Bao2021, Fatima2021}.
Most of these methods focused on identifying frame-by-frame in which an anomaly is occurred, but they are unrealized which road traffic participants can cause ego-vehicle leading into collision because of available annotation dataset only to detect anomaly on traffic video.

DADA-2000 \cite{Fang2021}, large-scale benchmark with $2000$ video sequences dataset having accident category of traffic accident video for both ego-vehicle involved and uninvolved, proposed annotation for risk accident for each frame.
Its annotation was focused to detect anomaly on the time accident occurred.
In our perspective, the time before and after accident occurrence has highly potential knowledge to learn the ego-motion during accident and damage probability occurring from ego-vehicle.
Near-miss is one type of accident and can be defined as a narrowly avoided accident that can be determined during or after accident times because there are several ways to avoid the accident such as fast response of either ego-driver or road traffic participant to take a brake.
However, there are no different between accident and near-miss on the time before accident happened, so that we re-define the definition of accident on DADA-2000 dataset together with near-miss and also extend start and end time of accident duration to precisely cover all ego-motions during incident.

Unlike previous works, our contribution is to classify all possible traffic risk accidents including near-miss to give more critical information for real-world driving assistance systems. %of which road participants involve ego-vehicle
Due to limited annotating video availability, we augment re-annotation DADA-2000 dataset using manipulating video style translation based on unsupervised image-to-image translation networks (UNIT) \cite{Mingyu2017} to increase the number of traffic risk accident videos and to generalize performance of video classification model on different types of conditions.

The aim of this research is to propose re-annotation DADA-2000 dataset which only focused on ego-vehicle involved by surrounding objects for both traffic accident and near-miss and to evaluate the performance of video classification model with and without conditional style translation.
For next, it can be useful for computer vision community to train their models to produce better traffic risk classification.
The proposed approach is derived from several existing works such as separable 3-dimensional convolutional neural network (S3D) \cite{Saining2017} and UNIT \cite{Mingyu2017}.
Traffic risk accident category is classified by S3D model using pre-trained model from Kinetics-400 while UNIT model is used to translate day-to-night and vice versa to augment our model for robust solution.

To summarize, we make the following \textbf{contributions}:
\begin{itemize}
\item We propose new consistency dataset re-annotation from DADA-2000 dataset focused on ego-vehicle involved by surrounding objects and re-define start and end time of traffic risk accident with additional near-miss category.
\item We propose the usage of UNIT model to translate day-to-night and otherwise to increase training, validating, and testing accident videos to generate robust classification model with small number of videos.
\item We show significantly improvement results of incident classification on cross validation analysis.
\end{itemize}

\section{Related Work}
This work is closely related to traffic accident dataset and classification in general and to translate day-to-night using image-to-image translation, which are discussed in the following subsections.

\subsection{Anomaly Detection Dataset in Egocentric Traffic Video}

\begin{figure}[t]
\begin{center}
   \includegraphics[width=\linewidth]{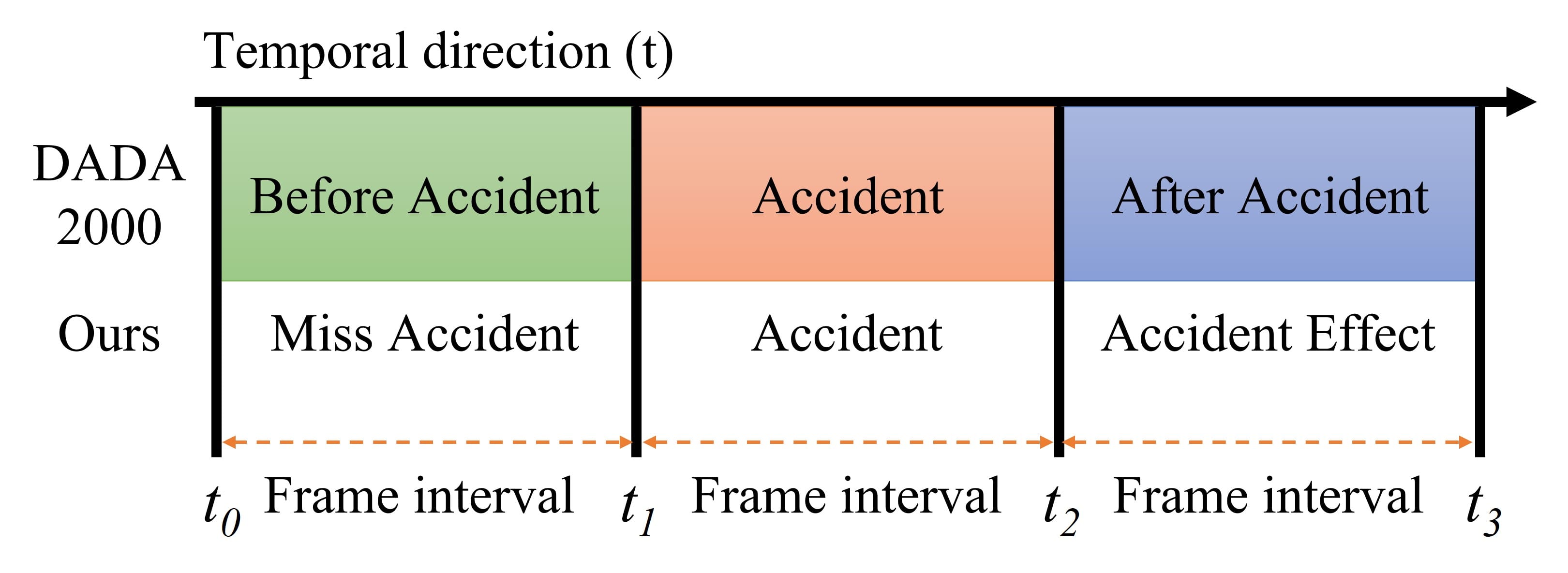}
\end{center}
   \caption{Illustration of clipped temporal window during accident.
   Our definition is that we have no way to recognize the accident during $t_0$ and $t_1$.
   There are several probabilities to put the ego-car into accident such as no response neither ego-car nor the object which leads into accident.
   }
\label{fig:figure21}
\end{figure}

Anomaly detection in egocentric traffic videos is attractive research because these events are very rare appearing on the nature.
Chan \etal \cite{Chan2017} proposed on-road accidents namely StreetAccident dataset with $620$ dash-cam video clips.
The temporal locations of accidents for each frame were manually annotated and $10$ frames remaining for each clip are defined as anomalous.
AnAn accident detection (A3D) dataset was proposed by Yao \etal \cite{yao2019} containing $1,500$ anomaly videos annotated by start and end times of abnormal event.
Similar with A3D, Fang \etal \cite{Fang2021} introduced the DADA-2000 dataset for driver attention prediction and $54$ kinds of accident categories based on road traffic participants.
From those accident categories, it can be classified into two sets based on affected from ego-vehicle: ego-vehicle involved and uninvolved, while Herzig \etal \cite{Herzig2019} extracted $803$ videos of accident and collision datasets from diverse driving dataset for heterogeneous multitask learning (BDD100K) \cite{yu2018}.
On the other side, a car crash dataset (CCD) dataset contains $1,500$ videos for traffic accident anticipation collected by Bao \etal \cite{bao2020}.
This dataset has accident causes and environmental attributes annotations.
Detection of traffic anomaly (DoTA) dataset \cite{Yao2022} annotates $4,677$ videos containing traffic risk accident of spatial (anomaly localization), temporal (anomaly duration), and categorical (anomaly types).
%Different from others, we are focusing ego-vehicle involved by surrounding objects such as pedestrian, cyclic, motorcycle, car, truck and self.
Different from others, we are focusing ego-vehicle involved by surrounding objects and near-miss incident is also added as class label from our definition shown in Fig. \ref{fig:figure21}.

\subsection{Traffic Accident Detection and Anticipation}
Future object localization (FOL) \cite{yao2019} predicted the future motion of surrounding objects and real ego-motion.
Its method used driver anomaly detection (DAD) \cite{chan2016} and A3D \cite{yao2019} datasets of dashboard-mounted camera and could be easily to classify both ego-vehicle involved or uninvolved.
On the other hand, Huang \etal \cite{Huang2020} combined both spatial and temporal stream of convolutional neural network (CNN) models that can perform object detection and tracking with classification model.
This method used CCTV video and reached over $86\%$ for F1 score.
Based on their results, research on accident detection using dashboard-mounted camera is more challenging problem due to relative motions between the ego-car and other vehicles.
Unlike CCTV video which has absolute motion patterns, relative motion patterns have the relationship between a pattern in the camera view cause camera position and ego-car movement.

On the other hand, traffic accident anticipation is still far from solved since the driver can still react after the time of prediction model result making more challenging problems.
Adaptive loss for early anticipation (AdaLEA) \cite{Suzuki2018} proposed early anticipation using an adaptive penalty weighted value.
Unlike AdaLEA model, spatio-temporal relational learning (UString) \cite{bao2020} learned spatial relations with graph convolutional networks (GCN) model by focusing recurrent neural network cell hidden states while deep reinforced accident anticipation with visual explanation (DRIVE) \cite{Bao2021} focused to integrate visually explanation for both top-down and bottom-up attention map by following human focusing before accident within a unified deep reinforcement learning framework.

\subsection{Image-to-Image Translation}
One of computer visions problems is image-to-image translation problem where an input image is transformed from source domain to target domain to be an output image.
Super-resolution, an example problem of image-to-image translation, is to map a low-resolution image to a high-resolution image.
In the unsupervised learning, there are two independent sets of images where no paired image between two independent sets.
The challenge for unsupervised learning is how an input image from source domain can be translated to an output image in target domain.

Recently, there has been increasing number of researchers focused on developing unpaired image-to-image translation to generate more realistic image result on target domain.
UNIT \cite{Mingyu2017} proposed the adversarial training objective interactions with a weight-sharing constraint to enforce a shared-latent space, and to generate corresponding images in two domains.
While tested, the various auto-encoders relate translated images with input images to the respective domains.
StarGAN \cite{Choi2017} introduced many-to-many translation based on their attribute labels.
The disadvantage using this architecture can be deterministic for a given input and domain.
StarGAN v2 \cite{Choi2019}, improving version of StarGAN \cite{Choi2017}, enabled a noise-to-latent mapping network to synthesize diverse results for the same domain.
Caused all of these methods depend on labels defined for classification, representations for image manipulation cannot be learned.
Contrastive learning based unsupervised image-to-image translation framework (CLUIT) \cite{Lee2021} recently proposed using contrastive learning through a discriminator.
However, the use of contrastive learning is only to replace the multi-task discriminator.

\section{Dataset Re-Annotation}

\begin{figure*}[t]
\begin{center}
   \includegraphics[width=\linewidth]{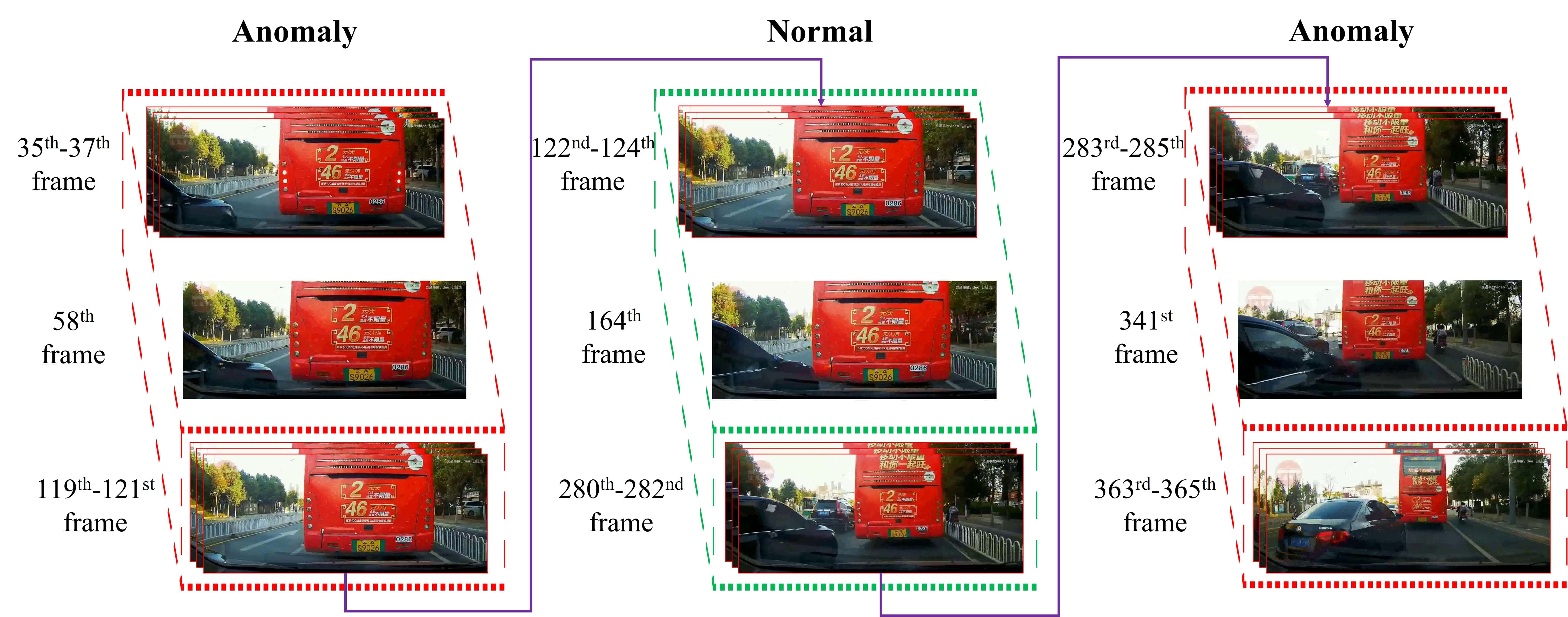}
\end{center}
   \caption{The visualization of DADA-2000 annotation.
   We can see that, DADA-2000 classify as normal on $122^{nd}$ to $282^{nd}$ frame while potential risk still can happen (it can clearly see on $164^{th}$ frame).
   }
\label{fig:figure31}
\end{figure*}

In our work, we use DADA-2000 dataset in which this dataset contains both ego-car involved ($520$ videos) and ego-car as observer ($493$ videos) because the number of shared annotation videos from Fang \etal \cite{Fang2021} is only $1013$ videos which is different from their reported.
Our work is to re-annotate DADA-2000 which only focuses on video of ego-car involved by surrounding objects with rich and precisely annotation including end-time of near-miss event.
It also improves the definition of start and end time of accident events with following annotation style of clipping video.
To know the potential risks of ego-vehicle, focusing the use of ego-car involved dataset can be the indicator to understand all potential events on the driver itself.
In Fig. \ref{fig:figure31}, we show example of inconsistency video annotation from DADA-2000 dataset.
Based on this figure, we can see that DADA-2000 annotated as normal on $122^{nd}$ to $282^{nd}$ frame.
In our perspective, the incident still can occur on this clipped duration.
So in our contribution, we re-annotate DADA-2000 dataset to improve consistency of its annotation.
Our annotation is available at \textbf{\url{https://github.com/jampang41/CST-S3D}}.

\subsection{Inconsistency DADA-2000 Annotation}

\begin{figure}[t]
\centerline{\subfloat[Uninvolved ego-car]{\includegraphics[width=0.45\textwidth]{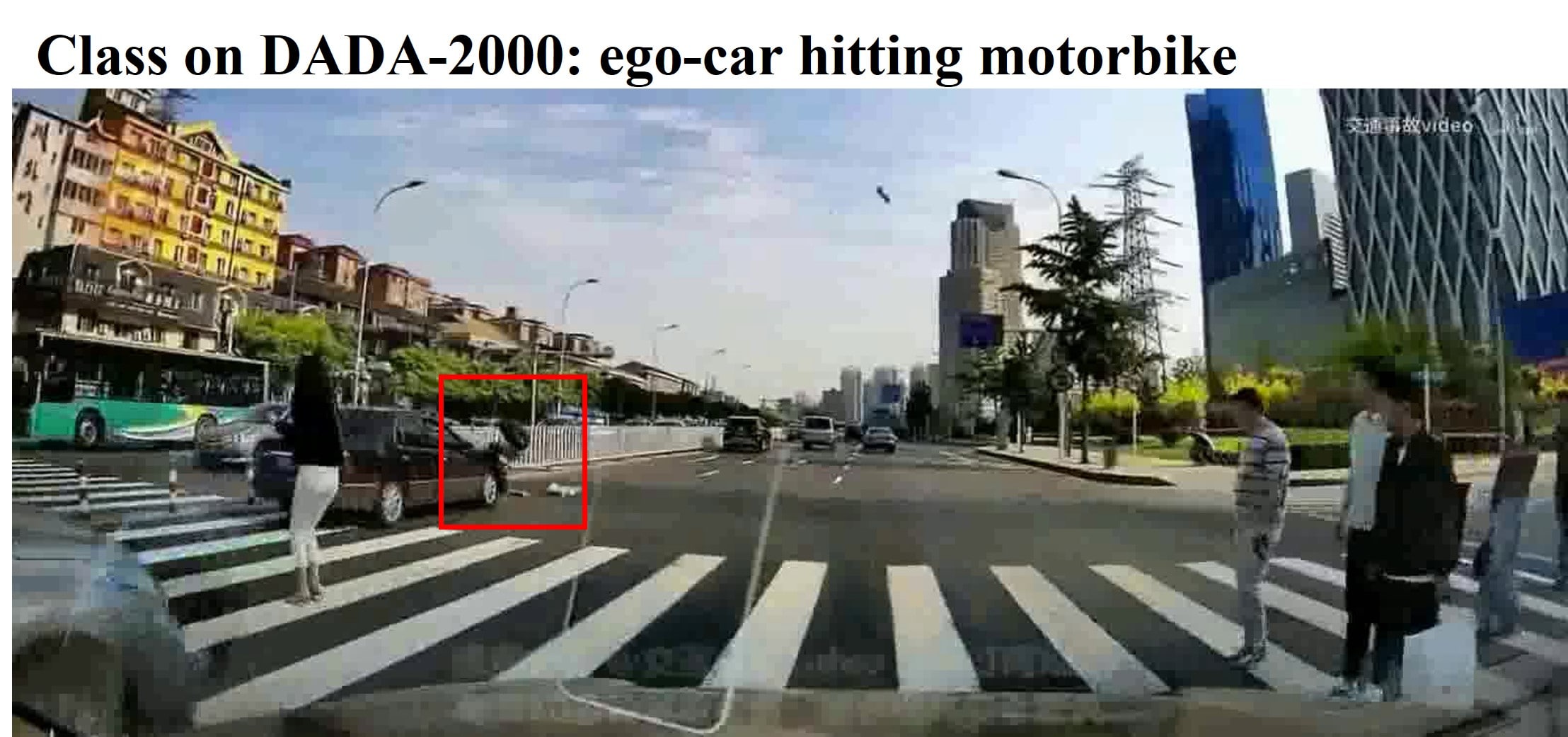}
\label{fig:figure321}}}
\vfil
\centerline{\subfloat[Wrong class annotation]{\includegraphics[width=0.45\textwidth]{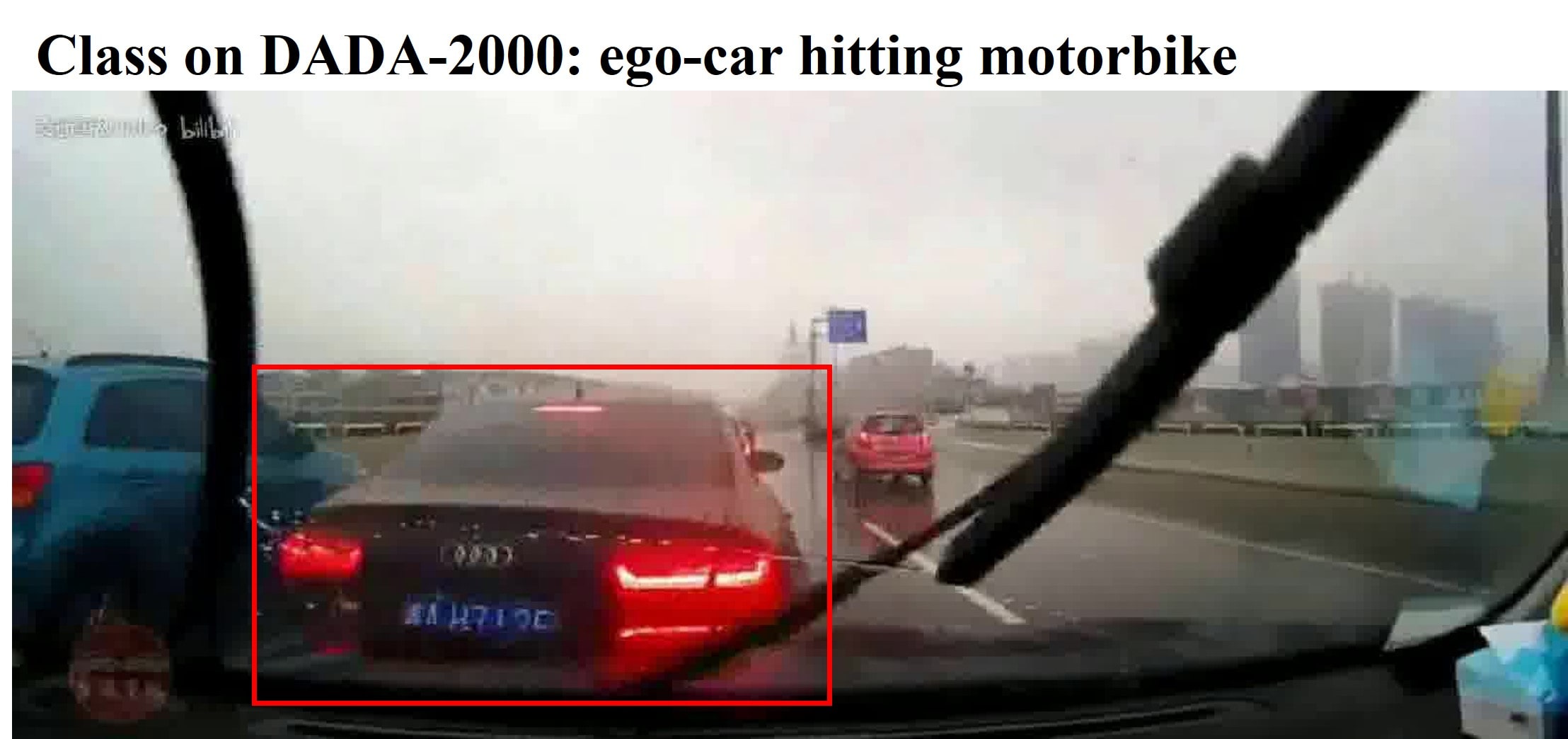}
\label{fig:figure322}}}
\vfil
\centerline{\subfloat[Video game dataset]{\includegraphics[width=0.45\textwidth]{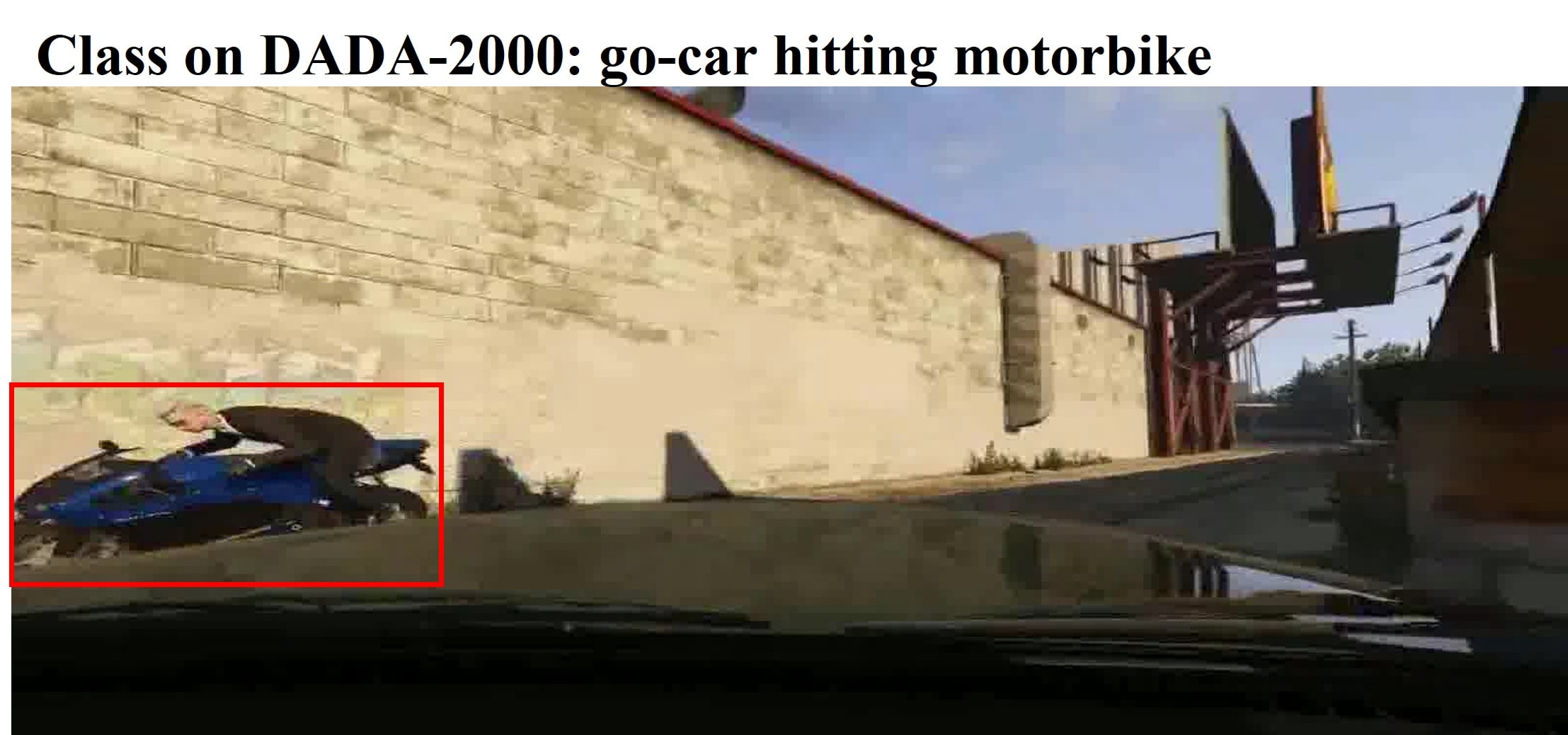}
\label{fig:figure323}}}
\vfil
\centerline{\subfloat[Normal annotation on risk incident frame]{\includegraphics[width=0.45\textwidth]{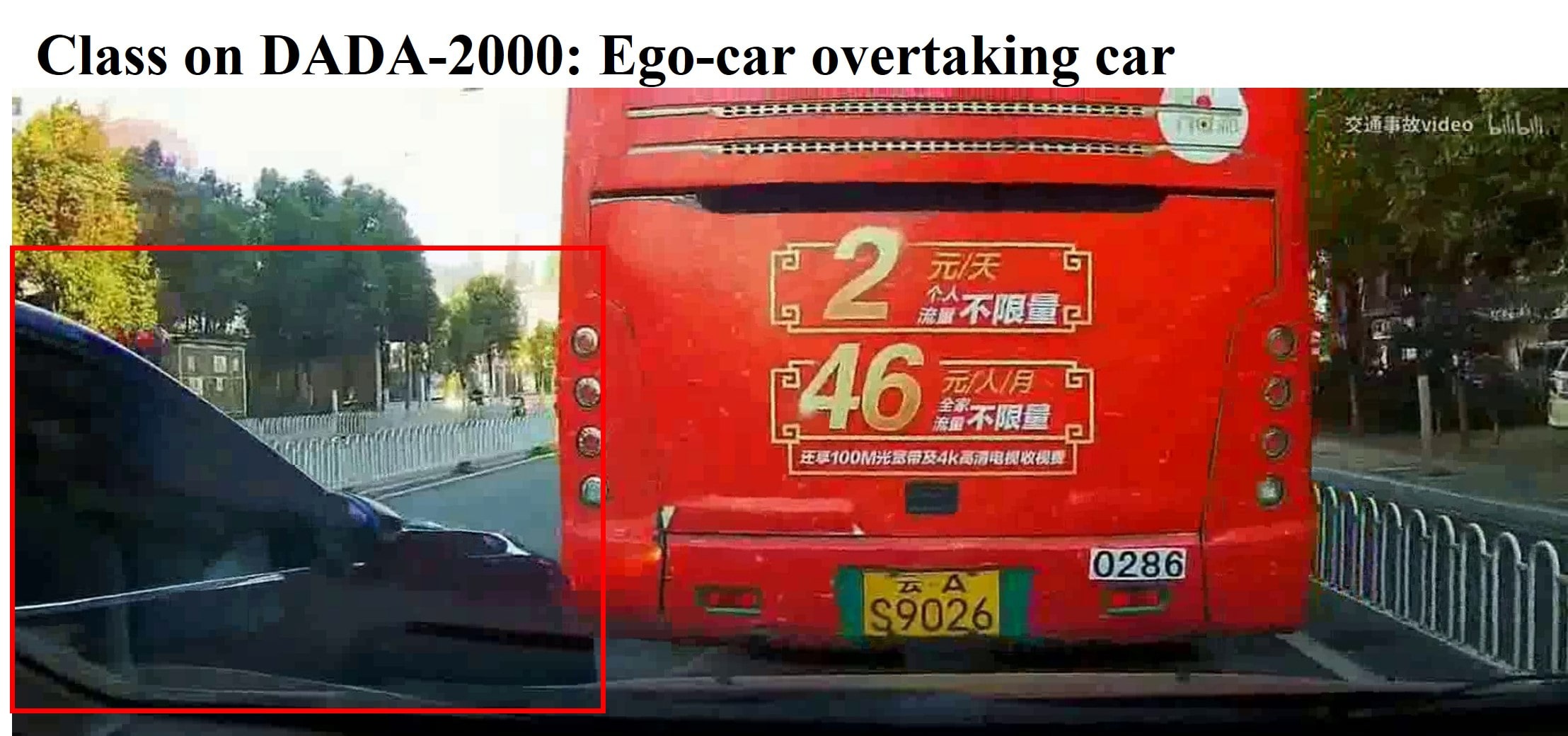}
\label{fig:figure324}}}
\caption{Inconsistency DADA-2000 annotation.
Red bounding box indicates road traffic participant that can lead ego-car into incident event. 
}
\label{fig:figure32}
\end{figure}

There are four different common inconsistent annotations of DADA-2000 dataset: uninvolved ego-car (Fig. \ref{fig:figure321}), wrong class annotation (Fig. \ref{fig:figure322}), video game dataset (Fig. \ref{fig:figure323}), and normal annotation on risk incident frame (Fig. \ref{fig:figure324}).
Based on Fig. \ref{fig:figure32}, the real event on dashboard-mounted video and its annotation are unmatched because inconsistency annotation can also cause inaccurate the classification model when it applies on real application.
In our annotation perspective, we remove the annotation containing uninvolved ego-car and video game dataset because our contribution is only focused to ego-car involved in real application.

\subsection{Dataset Annotation Technique}

\begin{table}[]
\centering
   \caption{different number of classes of re-annotation DADA-2000 dataset.
   }
\label{table:table31}
\begin{tabular}{@{}lll@{}}
\toprule
\multicolumn{1}{c}{4 classes} & \multicolumn{1}{c}{7 classes} & \multicolumn{1}{c}{\# 16 classes} \\ \midrule
1. Hitting pedestrian            & 1. Hitting pedestrian            & 1. Crossing pedestrian               \\
2. Hitting vehicles              & 2. Hitting cyclist               & 2. Hitting pedestrian                \\
3. Hitting obstacles             & 3. Hitting motorbike             & 3. Crossing cyclist                  \\
4. Normal                        & 4. Hitting truck                 & 4. Hitting cyclist                   \\ \cmidrule(r){1-1}
                              & 5. Hitting car                   & 5. Crossing motorbike                \\
                              & 6. Self-accident                 & 6. Hitting motorbike                 \\
                              & 7. Normal                        & 7. Crossing truck                    \\ \cmidrule(lr){2-2}
                              &                               & 8. Hitting truck                     \\
                              &                               & 9. Overtaking truck                  \\
                              &                               & 10. Crossing car                      \\
                              &                               & 11. Hitting car                       \\
                              &                               & 12. Overtaking car                    \\
                              &                               & 13. Hitting roadblocks                \\
                              &                               & 14. Hitting road facilities           \\
                              &                               & 15. Self-accident                     \\
                              &                               & 16. Normal                            \\ \cmidrule(l){3-3} 
\end{tabular}
\end{table}

\begin{figure}[t]
\begin{center}
   \includegraphics[width=\linewidth]{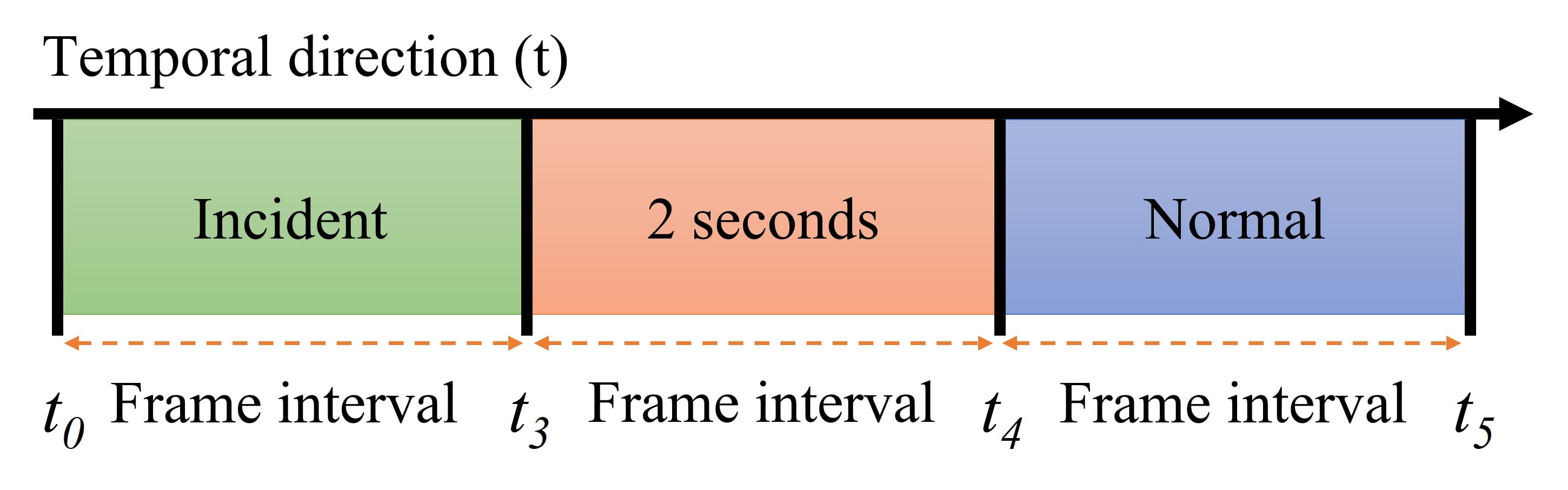}
\end{center}
   \caption{Illustration of clipped temporal window to clip incident and normal.
   To clearly separating incident and normal, we add $2$ seconds interval between $t_3$ and $t_4$.
   }
\label{fig:figure33}
\end{figure}

We carefully checked for each annotation and removed unusable video which is not belong to ego-car involved in real application.
Starting and finishing times $t_0$ and $t_3$ for each incident are also fixed by following the rule shown in Fig. \ref{fig:figure21}.
After fixing the annotation, each video is clipped from $t_0$ and $t_3$.
On the other hand, normal video is produced by normal case after incident shown on Fig. \ref{fig:figure33}.
To clearly separating incident and normal case, we added $2$ seconds after finishing time of incident $t_3$ to be starting time of normal case $t_4$ until the end of video $t_5$.
Given $\textbf{A}(n, \xi) = \{ A_1^\xi, A_2^\xi, A_3^\xi, \ldots, A_n^\xi\}$ is our annotation or ground truth labels of $n$ videos.
For detail information of our annotation or ground truth labels $\textbf{A}(n, \xi)$, we define as follow:

\begin{equation}\label{E:Equation31}
\begin{split}
\textbf{A}(n, \xi) &= \{ \textbf{V}(n), \textbf{P}(n, \xi), \mbox{\boldmath$ \mu$}(n)\}, \\
\textbf{V}(n) &= \{V_1, V_2, V_3, \ldots, V_n\}, \\
\textbf{P}(n, \xi) &= \{ P^n_1, P^n_2, P^n_3, \ldots, P^n_\xi\}, \\
\mbox{\boldmath$ \mu$}(n) &= \{ \mu_1, \mu_2, \mu_3, \ldots, \mu_n\},
\end{split}
\end{equation}

\noindent where $\xi$ is number of predicted classes and $\textbf{V}(n)$, $\textbf{P}(n, \xi)$, and $\mbox{\boldmath$ \mu$}(n)$ are annotated video, prediction class on re-annotation DADA-2000 dataset, and end time of near-miss incident, respectively.
%As a result, given $\textbf{P}(\xi) = \{ P_1, P_2, P_3, \ldots, P_\xi\}$ is prediction class on re-annotation DADA-2000 dataset where $\xi$ is number of predicted classes.
As a result, we generate prediction class $\textbf{P}^\vartheta(\xi)$ where $\vartheta$ is $3$ shown on Table \ref{table:table31} where each $\vartheta \in [1, 2, 3]$ equals $\xi \in [4, 7, 16]$, respectively.

%On the other hand, given near-miss label $\mu$ is added as the end time of 

\section{Methods}

\begin{figure}[t]
\begin{center}
   \includegraphics[width=\linewidth]{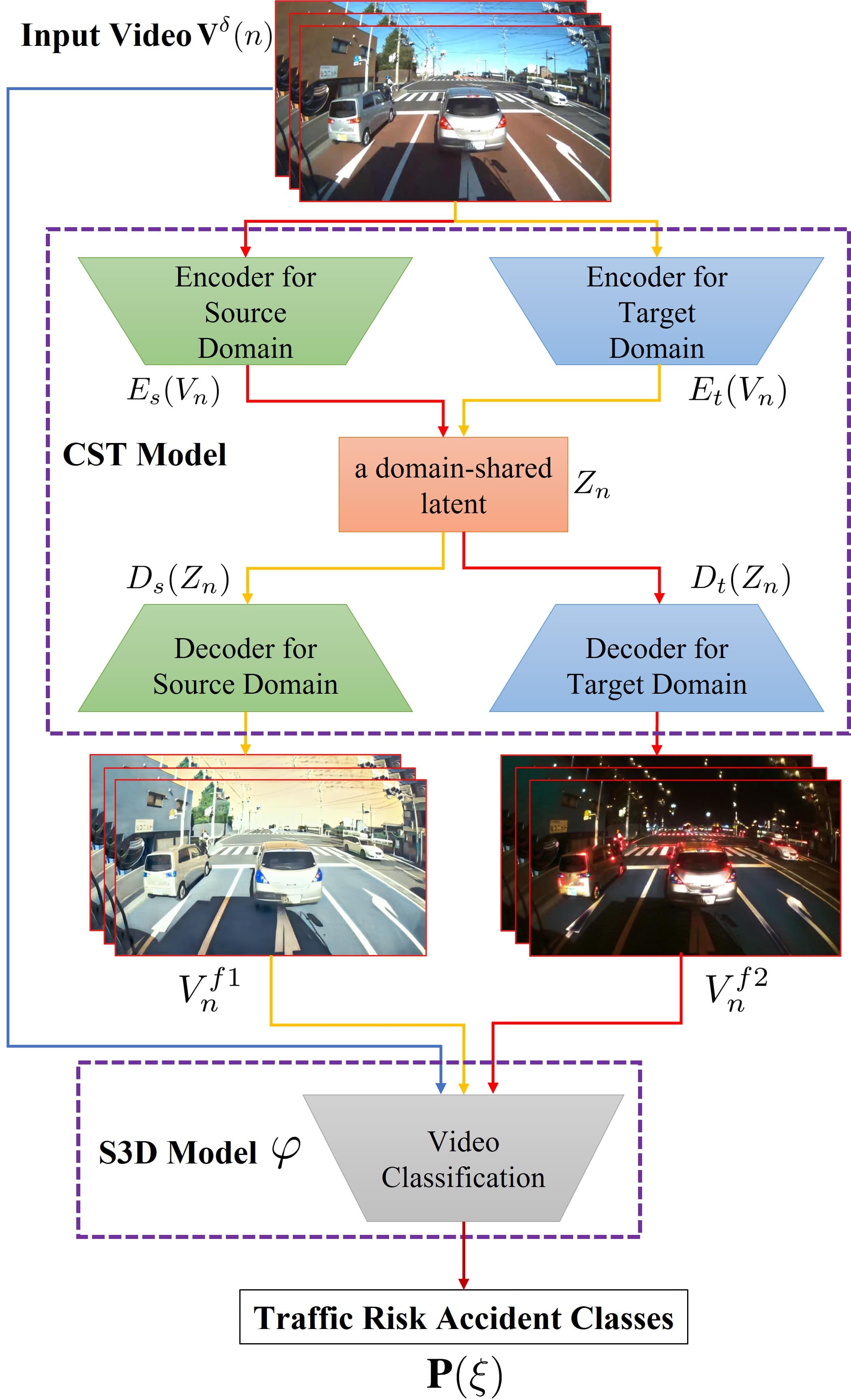}
\end{center}
   \caption{Diagram of the proposed CST-S3D architecture.
   The input video $\textbf{V}^{\delta}(n)$ is received and applied both encoder for source domain $E_s(V_n)$ and target domain $E_t(V_n)$ and reproduced $Z^{E_s}_n$ and $Z^{E_t}_n$, respectively.
   After that, $Z^{E_s}_n$ and $Z^{E_t}_n$ are inputted into decoder for target domain $D_t(Z^{E_s})$ and source domain $D_s(Z^{E_t}_n)$.
   As a result, we obtain $V^{f2}_n$ and $V^{f1}_n$ as fake videos produced by CST architecture.
 Its process is repeatedly until $n$ videos. Next, we train all videos using $\varphi$ to obtain final traffic accident classes result $\textbf{P}(\xi)$.
   }
\label{fig:figure41}
\end{figure}

In this section, we describe the detail explanation of the proposed method.
Our formulation is composed of two main components shown in Fig. \ref{fig:figure41}.
In the first component, we adopt conditional style translation (CST) differed from UNIT \cite{Mingyu2017} to increase number of accident videos for training, validating, and testing.
Another component uses S3D model \cite{Saining2017} to classify accident, near-miss and normal classes.
In order to provide some backgrounds and to formally introduce our approach, we start by providing our architecture of CST with S3D as video classification (CST-S3D) method.
The proposed CST-S3D model contains augmenting dataset using conditional style translation and classification modeling.
We then explain the implementation detail on our environment and performance of proposed system without videos produced by CST and cross validation dataset on real incident videos.
The system architecture of the proposed method is shown in Fig. \ref{fig:figure41} where frame-series of videos is an input of proposed system while the output is series of traffic risk accident classes.

\subsection{Conditional Style Translation (CST)}
Derived from UNIT \cite{Mingyu2017}, we generate two fake videos based on the shared-latent space assumption as illustrated in Fig. \ref{fig:figure41}.
Basically, the idea of CST architecture is that latent representations of a pair of corresponding images in two different domains into same latent space.
We are given $\textbf{V}^{\delta}(n) = \{V^{\delta}_1, V^{\delta}_2, V^{\delta}_3, \ldots, V^{\delta}_n\}$ as input original video $\delta$ of CST model where $n$ is the total number of input videos.
As shown in Fig. \ref{fig:figure41}, CST model architecture consists of four networks:
\begin{itemize}
\item $Z^{E_s}_n = E_s(V_n)$ is an encoder for source domain $s$ which is useful for extracting a domain-shared latent space $Z^{E_s}_n$ for each video on the source domain.
\item $Z^{E_t}_n = E_t(V_n)$ is an encoder for target domain $t$ which is useful for extracting a domain-shared latent space $Z^{E_t}_n$ for each video on the target domain.
\item $V^{D_s}_n = D_s(Z_n)$ is a decoder for source domain which is useful for generating a fake video $V^{D_s}_n$ from either source domain-shared latent space $Z^{E_s}_n$ or target domain-shared latent space $Z^{E_t}_n$.
\item $V^{D_t}_n = D_t(Z_n)$ is a decoder for target domain which is useful for generating a fake video $V^{D_t}_n$ from either source domain-shared latent space $Z^{E_s}_n$ or target domain-shared latent space $Z^{E_t}_n$.
\end{itemize}

\noindent As the result, we produce two different fake videos by following:

\begin{equation}\label{E:Equation41}
\begin{split}
V^{f1}_n = D_s(Z^{E_t}_n), \\
V^{f2}_n = D_t(Z^{E_s}_n).
\end{split}
\end{equation}

\noindent where $f1$ and $f2$ are fake video $1$ and $2$ produced by CST model.

%Differed from UNIT \cite{Mingyu2017}, we generate two fake videos using cross encoder-decoder of source and target domains.
%We also add two fake video types using to conditional style translation to increase number of dataset volume for training, testing, and validating.
%DST model are trained by following contribution of \cite{solesensei2019}.

\subsection{S3D}
There are many deep learning approaches using a large-scale data collection such as images and videos that has capability to resolve the limitation of current or usable methods with increased precision and accurateness.
In our application, we use S3D architecture which has a balance between speed and accuracy by building an effective and efficient video classification system through systematic exploration of critical network design choices.
We are given $\textbf{X}(n) = \{ \textbf{V}^{\delta}(n), \textbf{V}^{f1}(n), \textbf{V}^{f2}(n) \}$ as input video for classification model where $\textbf{V}^{f1}(n) = \{V^{f1}_1, V^{f1}_2, V^{f1}_3, \ldots, V^{f1}_n\}$ and $\textbf{V}^{f2}(n) = \{V^{f2}_1, V^{f2}_2, V^{f2}_3, \ldots, V^{f2}_n\}$.
Prediction class $\textbf{P}(\xi)$ is shown by following:

\begin{equation}\label{E:Equation42}
\begin{split}
\textbf{P}(\xi) = \varphi(\textbf{X}(n)).
\end{split}
\end{equation}

\noindent where $\varphi$ is the model followed by S3D architecture \cite{Saining2017}.

\section{Experiment}
In this section, we first explain implementation details to build proposed system.
We then describe evaluation approach to calculate accuracy to measure ability of proposed method and show quantitative evaluation on ablation study to discover the advantage of proposed method.

\subsection{Implementation Details}
In our implementation, we randomly separate the dataset into $70\%$, $20\%$ and $10\%$ of total images as training, testing and validating images without overlapping from each others and repeatedly apply random separating the dataset for each set $\textbf{V}^{\delta}(n)$, $\textbf{V}^{f1}(n)$, and $\textbf{V}^{f2}(n)$.
In our experiments, we utilized training and testing sets for evaluation performance.
During training process, we adopt the SGD optimizer with a mini-batch size of $2$.
The learning rate starts from $0.1$ and is divided by $10$ when reaching stable error with $0.9$ and $10^{-7}$ as momentum and weight decay, respectively.
The whole models are trained for up to $60$ epochs with MultiStepLR as scheduler.
We use four different image augmentations: random horizontal flip, random auto contrast, random gray scale, and random perspective with $0.5$ as probability.
Distortion scale is used on random perspective with $0.1$.
The entire training procedure takes about 24 hours using NVIDIA RTX$3090$ GPU with $24$ GB RAM.

\subsection{Evaluation Approach}
In our experiment, we use accuracy as measurement for evaluating performance of proposed method and state-of-the-art benchmark method.
Accuracy can be defined as the number of correctly predicted from the total number of entire data.
Multiclass data will be treated as if binarized under a one-vs-rest transformation.
Returned confusion matrices will be in the order of sorted unique labels in the union of ground truth $\textbf{A}(\xi) = \{0, 1\}$ and predicted class $\textbf{P}(\xi) = \{0, 1\}$.
In multilabel confusion matrix $\omega(\varphi)$, the count of true negatives (TN), false negatives (FN), true positives (TP) and false positives (FP) are $\omega^\varphi_{:A = 0,P = 0}$, $\omega^\varphi_{:A = 1,P = 0}$, $\omega^\varphi_{:A = 1,P = 1}$, and $\omega^\varphi_{:A = 0,P = 1}$, respectively.
Given accuracy $\varrho$ is defined as follow:

\begin{equation}\label{E:Equation51}
\begin{split}
\varrho(\varphi) = \dfrac{\omega^\varphi_{:A = 0,P = 0} + \omega^\varphi_{:A = 1,P = 1}}{\omega(\varphi)}.
\end{split}
\end{equation}

Best model $\varphi^*$ with maximum accuracy $\varrho$ is defined as:

\begin{equation}\label{E:Equation52}
\begin{split}
\varrho_\varphi = \operatorname*{arg\,max}_{c}(\mbox{\boldmath$ \varrho$}(\varphi_c))
\end{split}
\end{equation}

\noindent where $c \in [1, 9]$ is represented by the types of video classification method.

\subsection{Ablation Study}
\begin{table}[ht!]
\centering
   \caption{Accuracy result of six different models $\varphi_{c \in [1, 6]}$ where $\varphi_{c \in [1, 3]}$ are S3D model trained on original DADA-2000 annotation \cite{Fang2021} and $\varphi_{c \in [4, 6]}$ models are S3D model trained on  $\textbf{V}^{\delta}(n)$ with different size of video clipped for each model.
   As a result, the best result is shown in bold-text.
   }
\label{table:table51}
\begin{tabular}{lclc} 
\toprule
\multicolumn{1}{c}{Dataset}                                                               & Clipped Video   Size & $\varphi_c$ & Accuracy $\varrho$ (\%)  \\ 
\cmidrule{1-4}
\multirow{3}{*}{\begin{tabular}[c]{@{}l@{}}DADA-2000\\ (Baseline)\end{tabular}}           & 16                   & $\varphi_1$ & 31.08                    \\
                                                                                          & 32                   & $\varphi_2$ & 32.98                    \\
                                                                                          & 64                   & $\varphi_3$ & 31.04                    \\ 
\cmidrule{1-4}
\multirow{3}{*}{\begin{tabular}[c]{@{}l@{}}Re-annotation\\ DADA-2000 (ours)\end{tabular}} & 16                   & $\varphi_4$ & 37.22                    \\
                                                                                          & 32                   & $\varphi_5$ & 48.23                    \\
                                                                                          & 64          & $\mbox{\boldmath$ \varphi_6$}$ & \textbf{50.09 }          \\
\bottomrule
\end{tabular}
\end{table}

To prove that our annotation is corrected, we compare between original annotation of DADA-2000 dataset and our re-annotation.
We define six different models $\varphi_{c \in [1, 6]}$ where $\varphi_1$, $\varphi_2$, and $\varphi_3$ are S3D model trained on original DADA-2000 annotation using $16$, $32$, and $64$ frames as an input video clipped as baseline model while $\varphi_4$, $\varphi_5$, and $\varphi_6$ are S3D model trained on our annotation $\textbf{A}(\xi)$ using $16$, $32$, and $64$ frames as an input video clipped, respectively.
In this ablation study, we use $\vartheta$ and $\xi$ as $3$ and $16$, respectively.
The results of all testing videos are presented in Table \ref{table:table51}.
Based on this table, it proves that our models trained on our annotation $\varphi_4$, $\varphi_5$, and $\varphi_6$ have obvious result compared to the baseline by a large margin on accuracy measurement.
The proposed model $\varphi_6$ reaches $50.09\%$ with $17.11\%$ as positive margin from the best result of baseline model $\varphi_2$.
In this ablation study, we have demonstrated the effectiveness our annotation in application to incident classification and it can be widely used to train into various models.

\section{Results}

In this section, we investigate the effectiveness of adding CST architecture.
Firstly, we train S3D model on two different datasets: $\textbf{X}(n)$ and $\textbf{V}^{\delta}(n)$.
After that, we train and test both models with the same data transformation.
The database used in this experiment is re-annotation DADA-2000 dataset.
To evaluate the generalization capability, we trained CST-S3D on re-annotated DADA-2000 and tested on real incident videos.

\subsection{Adding CST architecture}
\begin{table}[ht!]
\centering
   \caption{The performance effect of CST model to produce fake videos $\textbf{V}^{f1}(n)$, and $\textbf{V}^{f2}(n)$ for enrich data training.
   We compare the S3D model trained on $\textbf{V}^{\delta}(n)$ namely $\varphi_{c \in [4, 6]}$ and $\textbf{X}(n)$ namely $\varphi_{c \in [7, 9]}$ with different size of video clipped for each model.
   As a result, the best accuracy for  are shown in bold-text.
   }
\label{table:table61}
\begin{tabular}{lclc} 
\toprule
\multicolumn{1}{c}{Methods} & Clipped Video   Size & $\varphi_c$ & Accuracy $\varrho$ (\%)  \\ 
\cmidrule{1-4}
\multirow{3}{*}{S3D}        & 16                   & $\varphi_4$ & 64.84                    \\
                            & 32                   & $\varphi_5$ & 66.38                    \\
                            & 64                   & $\varphi_6$ & 61.61                    \\ 
\cmidrule{1-4}
\multirow{3}{*}{CST+S3D}    & 16                   & $\varphi_7$ & 73.79                    \\
                            & 32          & $\mbox{\boldmath$ \varphi_8$}$ & \textbf{74.45}           \\
                            & 64                   & $\varphi_9$ & 67.06                    \\
\bottomrule
\end{tabular}
\end{table}

In this experiment, we evaluate performance of CST architecture by adding two fake videos $\textbf{V}^{f1}(n)$, and $\textbf{V}^{f2}(n)$ for additional training dataset on S3D method namely $\varphi_{c \in [7, 9]}$.
$\vartheta$ and $\xi$ as $2$ and $7$ are applied on this experiment to evaluate the effectiveness for different number of re-annotation classes.
By adding two fake videos $\textbf{V}^{f1}(n)$, and $\textbf{V}^{f2}(n)$, the accuracy of $\varphi_8$ reaches $74.45 \%$ with $8.07 \%$ as positive margin from the best result of the model $\varphi_5$ trained on $\textbf{V}^{\delta}(n)$.
Based on this experiment, CST architecture is effective to use as additional training set for increasing accuracy of video classification model.

\subsection{Cross Validation Analysis}

\begin{figure*}[ht!]
\begin{center}
   \includegraphics[width=\linewidth]{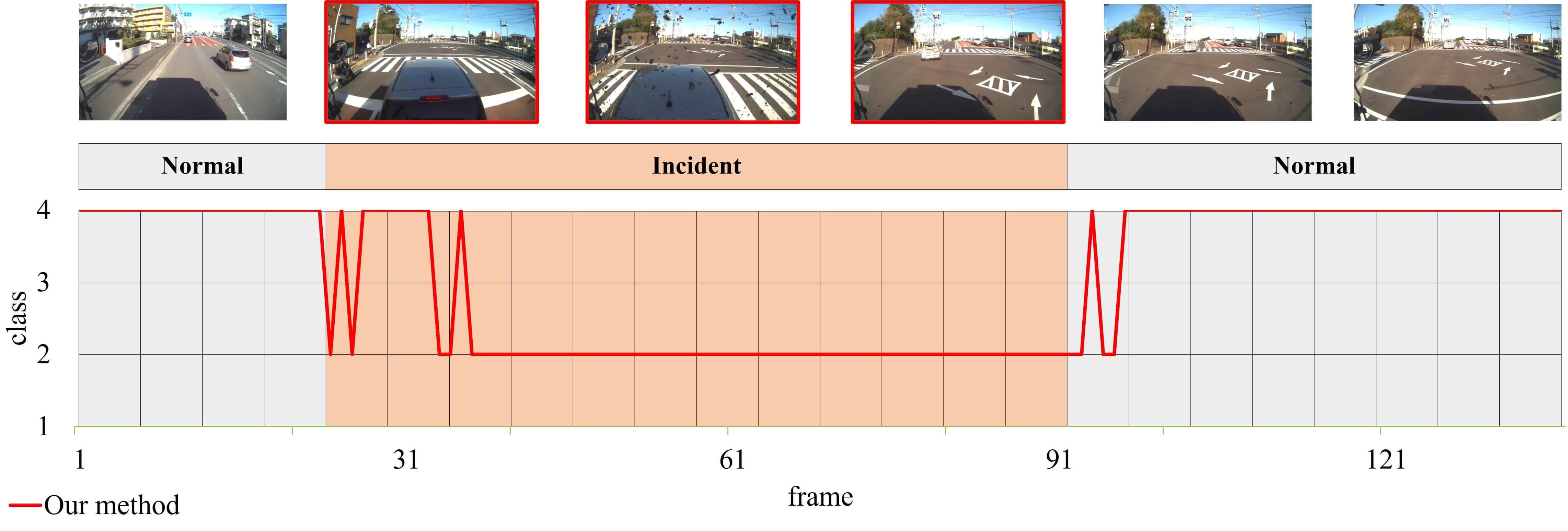}
\end{center}
   \caption{The visualization of cross validation analysis.
   We manually collect $39$ videos and test the model $\mbox{\boldmath$ \varphi_8$}$ on them.
   As the result, proposed system can consistently predict the incident class very well.
   }
\label{fig:figure61}
\end{figure*}

\begin{table}[ht!]
\centering
   \caption{The performance effect on cross validation analysis.
      We compare S3D method trained on DADA-2000 dataset annotation $\varphi_2$ and our contribution architecture CST-S3D $\varphi_8$ with $32$ frames as an input of video clipped.
   As a result, the best accuracy for  are shown in bold-text.
   }
\label{table:table62}
\begin{tabular}{lccc} 
\toprule
                                                                         & \begin{tabular}[c]{@{}c@{}}Number of\\ videos\end{tabular} & $\varphi_c$          & Accuracy $\varrho$ (\%)  \\ 
\midrule
DADA-2000           & 39                                                         & $\varphi_2$          & 48.72\%                                                            \\ 
\midrule
\begin{tabular}[c]{@{}l@{}}Re-annotation\\ DADA-2000 (ours)\end{tabular} & 39                                                         & $\mbox{\boldmath$ \varphi_8$}$ & \textbf{58.97\%}                                                   \\
\bottomrule
\end{tabular}
\end{table}

We manually collect $39$ videos to evaluate the effectiveness of proposed system.
In this experiment, we use $\vartheta$ and $\xi$ as $1$ and $4$, respectively.
In Table 5, the detail performance of proposed CST-S3D method $\varphi_8$ trained on $\textbf{X}(n)$ and benchmark S3D method \cite{Saining2017} $\varphi_2$ trained on original DADA-2000 dataset annotation.
Based on this table, $\varphi_8$ model is significantly improvement result comparing $\varphi_2$ model by $10.25 \%$ positive margin on accuracy and the best result is shown boldfaced.
For better visualization, we input previous $32$ frames to $\varphi_8$ model for predicting current classification class.
The example result of $\varphi_8$ model from $39$ videos is shown on Fig. \ref{fig:figure61}.
Based on this figure, the proposed method is demonstrated to be able classify well on cross data evaluation.
In our analysis, proposed method can classify the current classification class using previous information or frame and its method is valuable to apply in real environment.
By our experiment, our re-annotation on DADA-2000 dataset is enough to prove that our annotation is more consistency compared original annotation.

\section{Conclusion and Discussion}
Consistency dataset annotation is the one of important role to improve the performance of classification model to generalize classification model on real application.
Recent studies have shown that it is possible to build accident classification in entire of frames on video.
However, there is no agreement to increase performance of classification model in real accident video without increasing number of dataset and improvement of consistent dataset annotation.
In this paper, augmenting ego-vehicle for traffic near-miss and accident classification dataset using manipulating conditional style translation has been presented and demonstrated to be promising for improving the performance of classification model by generating fake video using CST model and improving DADA-2000 dataset annotation.
We have demonstrated proposed CST-S3D consistently performing well on the re-annotation DADA-2000 dataset and real accident video with higher accuracy comparing baseline model.
We expect our approach to open the door for future work and to go beyond to use our re-annotation DADA-2000 dataset and focus on generating on fake video on different seasons to enrich number of videos for training classification model.

\bibliographystyle{IEEEtran}
\bibliography{IEEEabrv,reference.bib}

\end{document}